# Entry-level guide to the use of large language models for medical research

Qiao Jin[1], Nicholas Wan[1], Robert Leaman[1], Shubo Tian[1], Zhizheng Wang[1], Yifan Yang[1], Zifeng Wang[2], Guangzhi Xiong[3], Po-Ting Lai[1], Qingqing Zhu[1], Benjamin Hou[1], Maame Sarfo-Gyamfi[1], Gongbo Zhang[4], Aidan Gilson[5], Balu Bhasuran[6], Zhe He[6], Aidong Zhang[3], Jimeng Sun[2], Chunhua Weng[4], Ronald M. Summers[7], Qingyu Chen[5], Yifan Peng[8], Zhiyong Lu[1]

**Author affiliations**

1. National Library of Medicine (NLM), National Institutes of Health (NIH).
2. Department of Computer Science, University of Illinois Urbana-Champaign.
3. Department of Computer Science, University of Virginia.
4. Department of Biomedical Informatics, Columbia University.
5. School of Medicine, Yale University.
6. School of Information, Florida State University.
7. Department of Radiology and Imaging Sciences, NIH Clinical Center.
8. Department of Population Health Sciences, Weill Cornell Medicine.

**Corresponding author**

Zhiyong Lu, Ph.D., FACMI, FIAHSI

Senior Investigator

National Library of Medicine

National Institutes of Health

8600 Rockville Pike

Bethesda, MD 20894, USA

E-mail: zhiyong.lu@nih.gov

## Abstract

Frontier large language models (LLMs), such as GPT-5, Claude 4.5, Gemini 3, Llama 4, and DeepSeek-R1, represent a transformative class of AI tools capable of revolutionizing various aspects of healthcare by generating human-like responses across diverse contexts and adapting to novel tasks following human instructions. Their potential application spans a broad range of medical tasks, such as clinical documentation, matching patients to clinical trials, and answering medical questions. In this paper, we propose an actionable guideline to help healthcare professionals more effectively and efficiently utilize LLMs in their work, along with a set of best practices. The overall workflow consists of several main phases, including formulating the task, choosing LLMs, prompt engineering, fine-tuning, and model deployment. We start with the discussion of critical considerations in identifying medical tasks that align with the core capabilities of LLMs and selecting models based on the selected task and data, performance requirements, and model interface. We then review the strategies, such as prompt engineering and fine-tuning, to adapt standard LLMs to specialized medical tasks. Deployment considerations, including regulatory compliance, ethical guidelines, and continuous monitoring for fairness and bias, are also discussed. By providing a structured step-by-step methodology, this entry-level tutorial aims to equip healthcare professionals with the tools necessary to effectively integrate LLMs into clinical practice, ensuring that these powerful technologies are applied in a safe, reliable, and impactful manner.

## Introduction

Large language models (LLMs), exemplified by GPT-5[1], Claude 4.5[2], Gemini 3[3], Llama 4[4], and DeepSeek-R1[5], are artificial intelligence (AI) models capable of generating human-like responses under various conversational contexts and adapting to novel tasks by following human instructions[6,7]. They have shown great promise in diverse biomedical and healthcare applications[8-12], such as question answering[13-16], clinical trial matching[17-19], clinical documentation[20-22], and multi-modal comprehension[23,24].

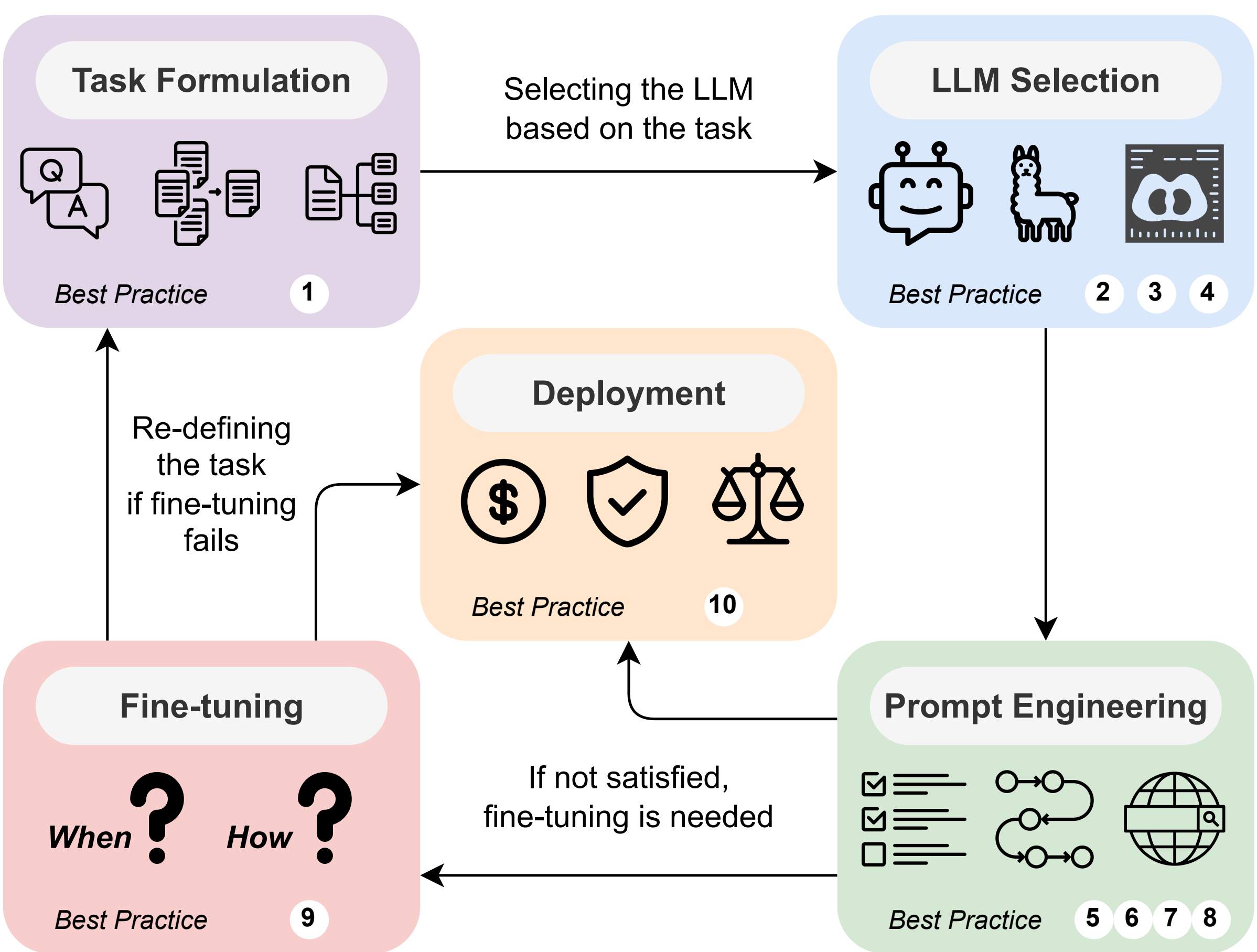

**Figure 1.** Overview of the proposed systematic approach to utilizing large language models in medicine. Users need to first formulate the medical task and select the LLM accordingly. Then, users can try different prompt engineering approaches with the selected LLM to solve the task. If the results are not satisfying, users can fine-tune the LLMs. After the method

development, users also need to consider various factors at the deployment stage. Corresponding best practices in Box 2 are also listed in each phase.

Despite the accelerated progress of LLMs in patient care and clinical practice, biomedical and health sciences research, and education[8,10,25-27], there is a noticeable shortage of practical, actionable guidelines for their application from bench to bedside and beyond. Much of the current use of LLMs, such as ad-hoc prompting with ChatGPT, is far from sufficient in medical tasks[28,29]. This gap can lead to both underutilization and misapplication of these technologies, potentially affecting patient outcomes. Our work aims to close this gap by providing a detailed, structured framework to guide the utilization and integration of LLMs into medical workflows (Figure 1). Specifically, this framework consists of task formulation, model selection, prompt engineering, fine-tuning, and deployment (See Box 1 for a glossary of technical terms). We further provide a set of best practices (Box 2) while supporting adherence to ethical use, evaluation metrics, and compliance standards. We hope this work will help equip healthcare professionals with the necessary knowledge to effectively leverage LLMs in their practices.

## Task Formulation

Adapting an abstractive medical need into one or more concrete tasks that can be addressed by an LLM requires users to first understand the core capabilities of LLMs, which we classify into five broad categories: (1) knowledge and reasoning, (2) summarization, (3) translation, (4) structurization, and (5) multi-modal data analysis. We recommend beginning by identifying the primary LLM capability relevant to your task. Once the task is formulated, aim to collect a diverse set of test cases (instances) that include input and output data elements for development (Box 2, Best Practice 1). We recommend collecting about 100 test instances, following several evaluation studies of LLMs in medicine[15,21,30].

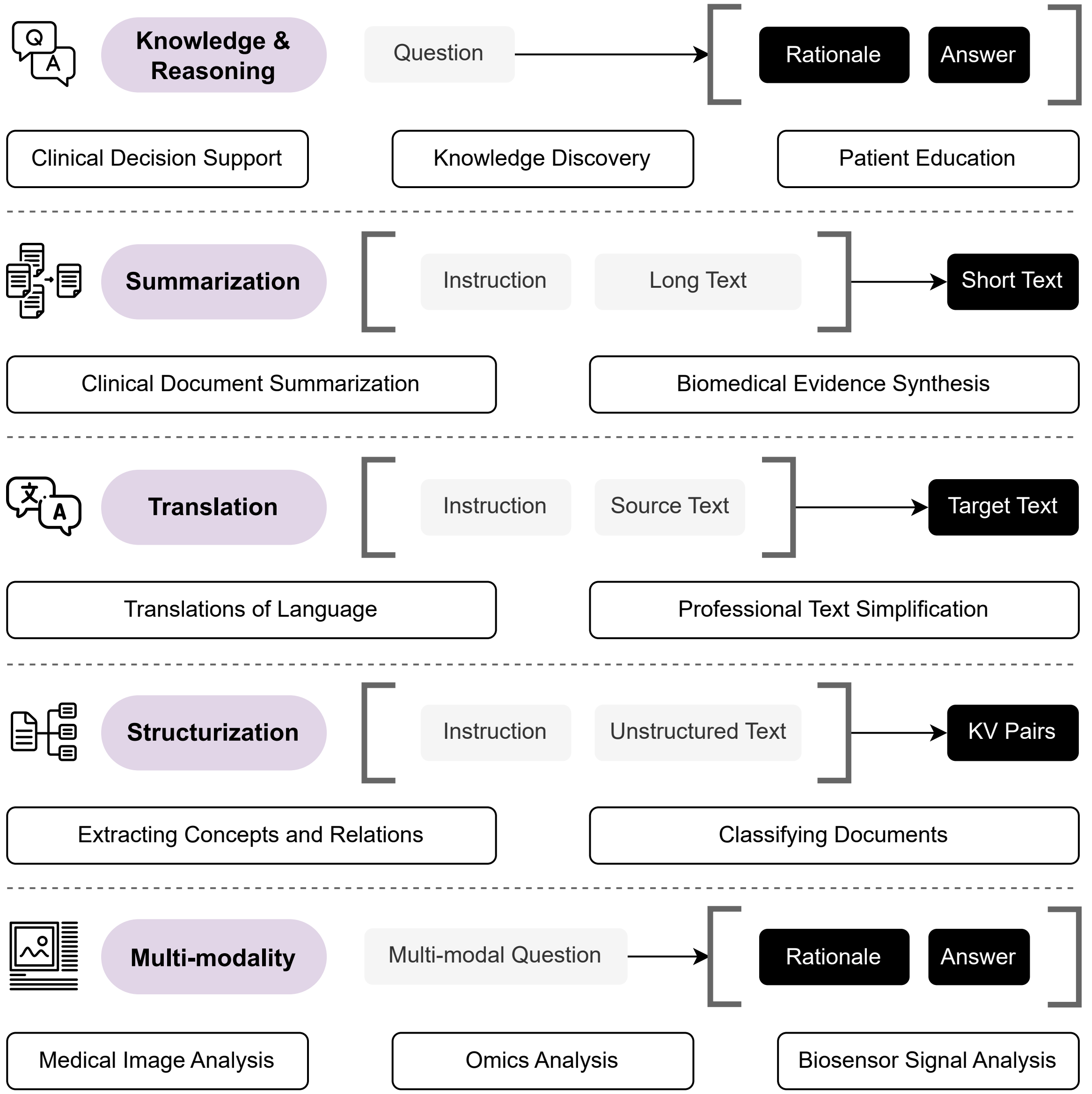

**Figure 2**. An overview of five common task formulations enabled by LLMs in medical research, with a set of examples. LLMs can answer questions using their domain knowledge and reasoning capabilities. The summarization task shortens long texts into concise summaries. The translation task transforms the source text to the target text in different language styles. The structurization task converts unstructured texts into structured key-value (KV) pairs. LLMs can also be used to support multi-modal data analysis such as interpreting medical images.

**Knowledge and reasoning**

LLMs can use the medical knowledge encoded in their parameters to perform domain-specific reasoning given different contextual information[14,15,31-34]. This capability can enable a variety of medical applications, such as answering medical questions[35], clinical decision support[36], and matching patients to clinical trials[17-19]. Task instances usually include question (input), explanation (reasoning and context output), and short answer (output). When formulating a reasoning task, users can quickly evaluate short answers (e.g., yes/no) between LLM predictions and ground truth and proceed to in-depth analyses of explanations if short answers are satisfying[30].

**Summarization**

LLMs can summarize complex documents into concise paragraphs or sentences. Summarization tasks in biomedicine primarily fall into two categories: (1) summarizing long clinical notes into shorter texts, such as generating the progress notes and discharge summaries[20-22]; (2) summarizing medical literature for evidence synthesis, such as generating the systematic reviews given a list of clinical studies[37-39]. These labor-intensive tasks can be potentially streamlined by LLMs. For a summarization task, instances include instruction (input), original text (input), and summarized text (output). Users may leverage metrics like BLEU[40], ROUGE[41], and BERTScore[42] to compare the generated texts and the reference summaries. However, it should be noted that automatic metrics do not always correlate well with the gold-standard human judgments[43].

**Translation**

LLMs also have the capability to translate text, not only between different languages but also between writing styles appropriate for different audiences. This ability can enable applications such as sharing medical knowledge across different language demographics[44], supporting medical education[45,46], and facilitating communication with patients[47]. A translation task, similar to a summarization task, includes instances made of instruction (input), source text (input), and target text (output) and is evaluated similarly to a summarization task.

### Structurization

LLMs can be leveraged to convert free-text input into structured outputs such as a list of key-value pairs. Medical structurization problems include classifying an input text into controlled vocabularies like the diagnosis-related groups[48] and extracting biomedical concepts as well as their relationships (e.g., variant-*causing*-disease) from unstructured text[49]. Structurization instances include task instruction (input), source text (input), and a list of extracted concepts and relations in structured forms (output). The evaluations are made to match the output with the reference answers. As such, the performance is often typically measured by precision, recall, and F1 score, etc.

### Multi-modality

Multi-modal LLMs like GPT-5 can analyze and integrate diverse data types such as text, images, audio, and even genomic information, potentially serving as a generalist medical AI[23]. For example, these models can support clinical tasks, such as generating radiology reports and guiding clinical decisions[50-52] based on real-world multimodal patient data. The multi-modal task instances and evaluations are similar to those of the knowledge and reasoning tasks, except that the input question typically contains data in multiple modalities (e.g., past medical history in EHR with current imaging results).

## Large Language Model Selection

Users should choose an appropriate LLM based on their task characteristics. There are a wide variety of LLMs, including proprietary models such as GPT-5[1], Claude Opus 4.5[2], and Gemini 3[3], and open-source models such as Llama 4[4], Mistral[53], Qwen3[54], and DeepSeek-R1[5]. They can range in size from several billion parameters (e.g., Qwen3-4B) to hundreds of billion parameters (e.g., DeepSeek-R1-671B). Typically, larger models exhibit more proficient responses[55]. Some models are trained for more general applications, while others like PMC-LLaMA[56] and MEDITRON[57] are fine-tuned for specific domains or applications. Newer thinking models, such as DeepSeek-R1 and OpenAI o-series LLMs, are often incentivized for improved reasoning capabilities and have achieved state-of-the-art performance on medical benchmarks[5,58,59]. When choosing the LLM, users should primarily consider three key factors in their needs: Task and Data, Performance Requirements, and Model Interface. Figure 3 shows an overview of the LLM selection considerations discussed in this section, and Table 1 shows the characteristics of commonly used LLMs.

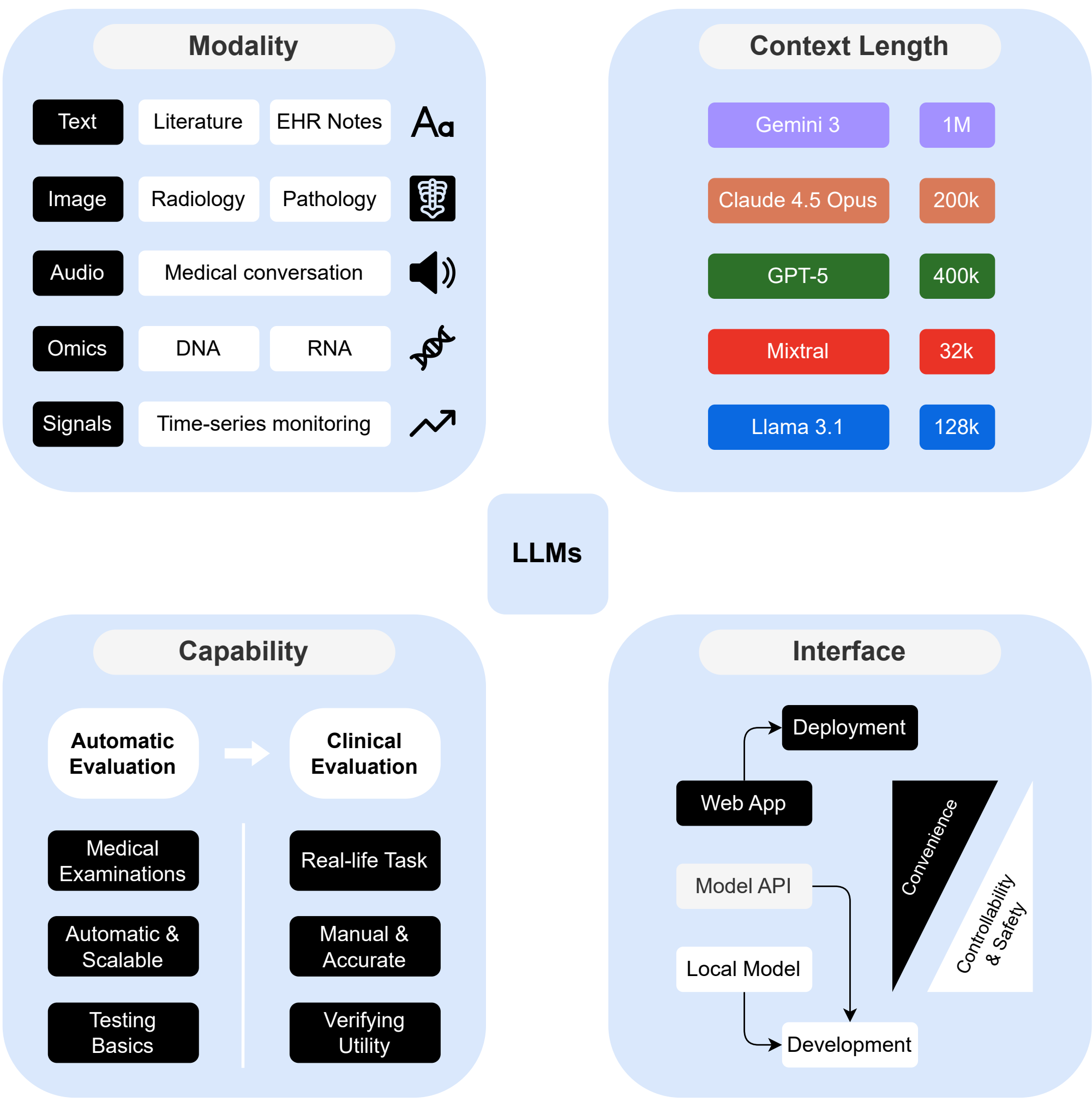

**Figure 3**. Considerations for choosing the LLMs. Users need to choose LLMs that can handle the modality and context length of the selected task. They also need to understand the capabilities of LLMs in the medical task. The gold standards come from manual evaluation of the model's behavior on real clinical tasks, but this is expensive and time-consuming. Models might be screened for basic medical capabilities with automatic evaluations on medical examinations first, and only models that pass the medical examinations are suitable for clinical evaluation. During the development phase, users need to use the model APIs or local models for better controllability and safety features. Web applications such as online chatbots are suitable for deployment to reach more users. EHR: Electronic medical records. MCQ: multiple-choice questions.

**Task and Data**

The first and most critical factor to consider is the nature of the data the user is working with and the specific task to perform. Ensuring that the chosen model aligns with the data type and task requirements is foundational to successful LLM implementation. When working with sensitive patient data, it is important to ensure privacy and compliance with regulations such as the Health Insurance Portability and Accountability Act (HIPAA)[60]. Proprietary models accessed through APIs like OpenAI's GPT-5 are typically not HIPAA-compliant, so they should not be used for patient data. In contrast, certain cloud service providers such as Azure and Anthropic provide HIPAA-compliant access to LLMs, which could be potentially used for sensitive data. Alternatively, local models such as Llama or Mistral can be used for enhanced control over privacy and security when processing sensitive medical information.

The diversity of healthcare tasks necessitates processing various data modalities in addition to free texts. Radiology and pathology applications, for example, may require models that can interpret and generate insights from 2D or 3D medical images, which requires models beyond text-only LLMs like o3-mini and Llama 3. Medical conversations produce audio data, which can be transcribed into text for processing by text-based LLMs. Genomics data, including DNA sequences and RNA expressions, require the knowledge of omics data interpretation[61]. Similarly, time-series data, such as monitoring vital signs, need models that can analyze long temporal patterns in structured EHR. While both genomics and time-series data can be represented as free-text, it remains unclear whether general LLMs can effectively handle such inputs without further training or adaptation. Users should determine what data modalities are essential to their task and select an LLM that can support such modalities (Box 2, Best Practice 3).

For tasks that involve large inputs, the length of the input data that the model can handle is crucial. Users must understand the length distribution of their datasets and choose an LLM with a context window that can accommodate the input data (Box 2, Best Practice 3). The title and abstract of a PubMed article, for instance, consist of roughly 250 words, or about 300-400 tokens (model input units). As demonstrated in the online tutorial, one token is about 0.8 words as text is often tokenized into subwords and individual characters. Some open-source models like Llama 3.0 have a limited context window (the longest input prompt it can process) of 8,000 tokens, which is about 20 abstracts. In contrast, long-context LLMs like GPT-5 (400k tokens context window), Claude 4.5 Opus (200k tokens), and Gemini 3.0 (1M tokens) can process approximately 1,600, 800, and 2,500 abstracts, respectively. Selecting LLMs with appropriate context windows ensures efficient processing of long input, but it should be noted that issues such as lost-in-the-middle[62] (where the model fails to utilize information in the middle of the prompt) also appear in long-context LLMs.

**Performance Requirements**

The model's medical capabilities are one of the most critical factors to consider when selecting LLMs. Typically, greater capabilities come with larger model sizes, which require more resources for development and customization. Conversely, smaller LLMs may not perform as well as their larger counterparts, but they are often more sustainable and less costly. While LLM capabilities vary across different model sizes, they are also affected by the target applications. For example, LLMs are better for tasks that require medical knowledge and clinical reasoning, but do not often outperform fine-tuned BERT models[63] in simpler tasks such as structurization[64].

There are two main approaches to evaluating the medical capabilities of LLMs: automatic evaluations, which use standardized benchmarks, and clinical evaluations, which assess performance in real-world healthcare contexts. Medical examinations in the form of multiple-choice questions (MCQ), such as MedQA-USMLE[65], PubMedQA[66], MedMCQA[67], have been commonly used to evaluate the knowledge and reasoning capabilities of LLMs[15]. These benchmarks are scalable and can serve as a starting point for medical AI evaluation. However, higher scores on these datasets do not necessarily translate to better clinical utility, as there are no choices provided in real-life applications and many studies have shown that MCQ performance does not correlate with real-world task utility[30,68,69]. Therefore, open-ended evaluations such as rubric-based HealthBench[70] and preference-based Chatbot Arena[71] represent promising future directions for exploration. After a model passes initial screening via automatic evaluations, it must be further assessed for clinical utility. This involves rigorous testing, such as randomized controlled trials, to ensure the model's outputs are trustworthy and beneficial in real-world healthcare settings.

In summary, automatic evaluation can be used to screen LLMs for basic medical capability in a scalable way, while clinical evaluation can provide a gold standard relevant to patient care at the cost of greater human effort. When selecting LLMs, we recommend that users choose LLMs based on clinically evaluated results. However, clinical evaluation is challenging and may not be readily available. In such scenarios, users should consider using automatic evaluation results to guide the selection of LLMs for further clinical evaluations. (Box 2, Best Practice 4)

**Model Interface**

Once the LLM(s) has been chosen, the users also need to select a point of access to the LLMs based on their needs. Broadly, there are three ways to access LLMs: web applications, model application programming interfaces (APIs), and locally hosted implementations. Web applications, such as ChatGPT, are inexpensive and easy to use compared to APIs and local models; however, they do not provide interfaces that allow flexible control of model behavior and large-scale evaluations. In addition, most LLM web applications do not have clear compliance with standards such as HIPAA, which further raises security issues when dealing with sensitive patient data. Consequently, we do not recommend using web applications during the development phase. In contrast, model APIs are controlled points of access via the web that provide an interface to proprietary models such as GPT-5 and open-source models like Llama 4. They are typically easier to use than implementing local LLMs, and some of the model APIs provide HIPAA-compliant services. Lastly, locally hosted LLM implementations are often derived from open-source models. In general, local models provide greater control and privacy[72]. For instance, accessing specific parameters and getting the raw prediction of the next token distribution is possible in a local model, but often not with a model API or via Web applications. Overall, while cutting-edge proprietary LLMs often deliver better performance in general tasks, users may have less control over customization, privacy, and safety. Conversely, open-source LLMs allow for greater customization and security but have higher requirements of hardware (such as computing resources) and technical expertise for both the development and the deployment phases.

Given the variety of LLMs available for medical applications, LLM selection requires careful considerations. While the supported data modalities and context windows are hard constraints, users must navigate trade-offs between model interface and medical capability based on their specific needs. For example, users may want to select proprietary LLMs of larger size when general capability has high priority. In contrast, users may opt to use local models when customizability is a concern. Ultimately, users must examine their task and select one (or more) LLM that satisfies their priorities.

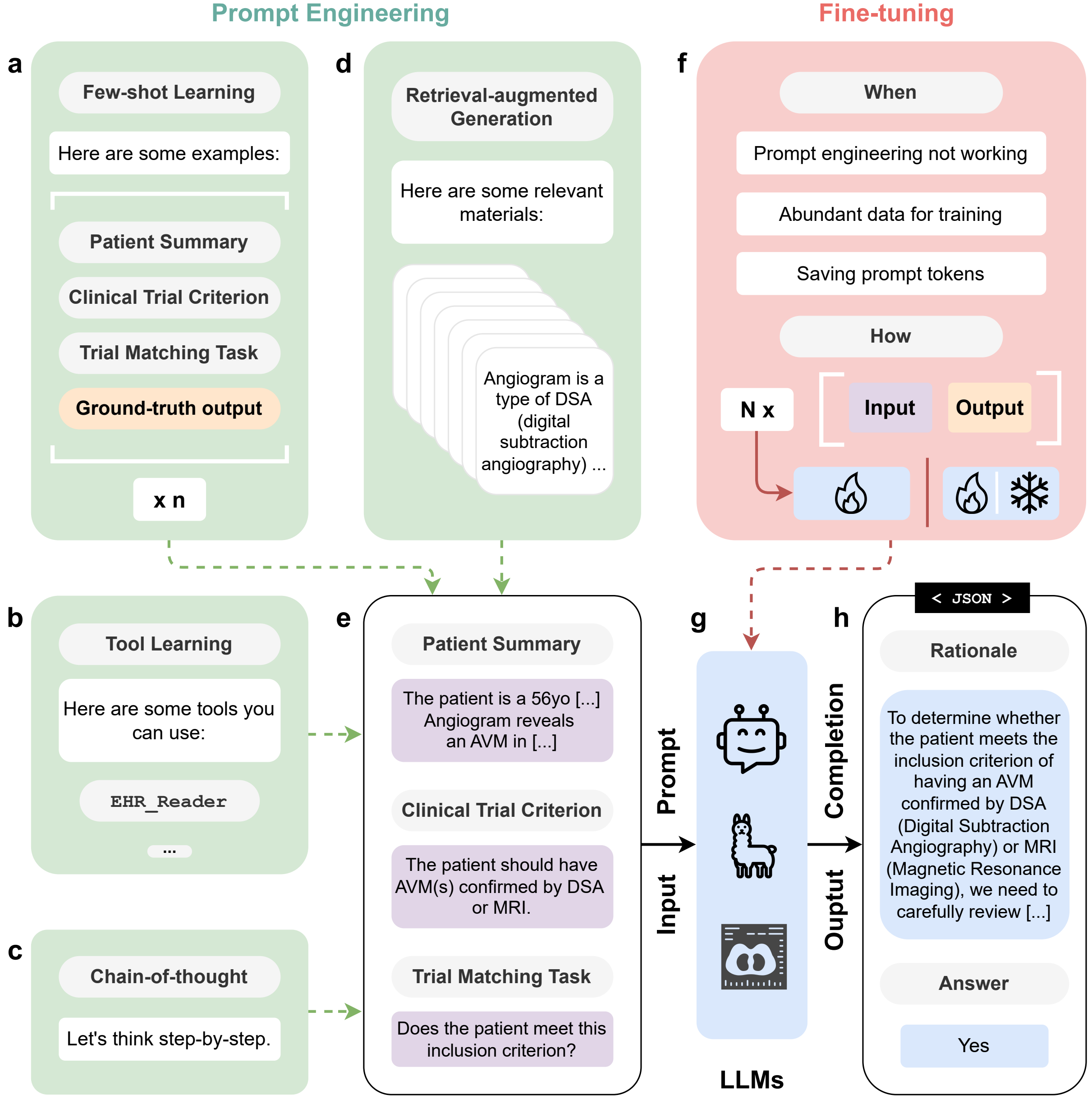

**Figure 4**. An overview of prompt engineering and fine-tuning techniques. **a**, Task examples are shown to the model in few-shot learning (FSL). **b**, Tool learning provides the model with access to external tools like database utilities. **c**, Chain-of-thought (CoT) prompting instructs the model to generate step-by-step rationale. **d**, Retrieval-augmented generation (RAG) provides relevant materials to solve the task. **e**, The patient-to-trial matching task where the patient summary and the clinical trial eligibility criterion are given. **f**, An overview of fine-tuning, including when and how to perform it. **g**, The inputs to LLMs are known as "prompts", and their outputs are "completions". **h**, An example output of LLMs that contain the CoT rationale as well as the short answer, organized in the JSON format. n denotes the number of shots for few-shot learning, and N denotes the number of instances for fine-tuning.

## Prompt engineering

Harnessing LLM capabilities for application to a specific task requires careful consideration of the prompt (input content) given to the model. Prompt engineering is the process of designing and optimizing prompts to effectively guide LLMs in generating accurate and coherent responses[73]. Prompts can vary from one simple instruction to many documents retrieved by a search system, allowing users to elicit a variety of behaviors without the need to modify the parameters of LLM. In general, more complex tasks typically require more sophisticated prompts. Prompts can be manually designed with iterative feedback or automatically optimized by LLMs themselves using frameworks like DSPy[74] and TextGrad[75], both of which enable programmatic prompt optimization by automatically adjusting and refining prompts based on performance feedback. Figure 4 shows a concrete task example of clinical trial matching, where a simple prompt that merely describes the patient and lists the clinical trial criteria (shown in Fig. 4e) might not work well. As such, prompt engineering and fine-tuning methods should be used. Table 2 shows resource requirements, advantages, disadvantages, and use cases of different approaches. Users need to understand the pros and cons of each technique. During prompt engineering, users can analyze the errors and choose a technique that mitigates them. For example, if an error is due to the lack of medical knowledge, retrieval-augmented generation can be used to provide relevant information for the model to better perform the task. Some detailed case studies are also listed in Table 3.

### Few-shot learning (FSL)

As shown in Fig. 4a, FSL includes a few examples (i.e., "shots") within the prompt to better specify the task for the model[6]. Each demonstration example should include both the input and the desired output. In the case of patient-to-trial matching, the input contains patient information, clinical trial criterion, and the task instruction; the output contains the rationale and the criterion-level eligibility label. Zero, one, and more than one examples (e.g., five) are respectively denoted as zero-shot, one-shot, and few-shot learning and are the most commonly used in practice. The examples should be as representative and diverse as possible. For example, demonstrations of all potential labels (e.g., disease subtypes) should be shown for a classification task. Another useful approach to few-shot learning is to generate examples dynamically, based on semantic similarity to the instances being predicted[14]. We recommend starting from zero-shot learning and incrementally adding examples to increase performance or deal with edge cases (Box 2, Best Practice 5).

**Chain-of-thought (CoT) prompting**

As shown in Fig. 4c, CoT prompting involves designing prompts that lead the model through a step-by-step reasoning process[76]. For example, providing the explanation of the patient-criterion relation helps the users efficiently verify the LLM-predicted eligibility labels. One can simply add "Let's think step-by-step" to the end of the input for CoT prompting. This technique is particularly useful in complex medical decision-making tasks, where an explanation of reasoning can improve model performance and aid clinicians in understanding and verifying AI-generated advice. We recommend using CoT prompting as the default as it improves the explainability of AI responses and potentially the performance as well (Box 2, Best Practice 6).

**Retrieval-augmented generation (RAG)**

In RAG (Fig. 4d), a search engine retrieves relevant documents, such as scientific articles, to be included in the prompt, allowing the model to better solve knowledge-intensive tasks such as answering questions[77] (Box 2, Best Practice 7). By grounding the LLMs to respond based on the provided relevant textual snippets, RAG can potentially reduce hallucinations[78,79] (the generation of incorrect content or references[80]) and improve upon outdated knowledge encoded in large language models. Both issues can be especially concerning in medical applications given the potential impact on patient safety and clinical decision-making. As an example of RAG, LLMs can get access to the definition of certain medical concepts with RAG to better classify patient eligibility in the application of patient-to-trial matching. We recommend using high-quality domain-specific corpora, such as systematic reviews, medical textbooks, and clinical guidelines, for RAG systems in medicine[79,81,82]. Commonly used resources include PubMed[83,84] with over 39 million freely available abstracts, and PubMed Central with over 7 million open access full-text articles for text and data mining[85]. Several generative search engines that adopt RAG to summarize the retrieval results are also available in biomedicine[84].

**Tool learning**

Certain medical tasks require the use of domain-specific tools such as database utilities[86,87] or medical calculators[88]. If these tools are implemented as application programming interfaces (APIs), LLMs can utilize them through a function calling mechanism (Box 2, Best Practice 7). In the example case of clinical trial matching, LLMs can be provided with tools for reading raw electronic health records to capture detailed information that might be missing from a patient summary[89] (shown in Fig. 4c).

**Setting the temperature**

Besides the prompt, LLMs also require a temperature parameter that controls the amount of randomness when generating the response. Lower temperatures result in a more consistent output, while higher temperatures lead to more creative responses. Temperature can usually be set via API or local parameterization, but usually not via Web app. We recommend that users start with a temperature of 0 to get deterministic results for reproducibility and only consider increasing the temperature to get diverse responses for purposes such as ensembling[90] (Box 2, Best Practice 8).

### Additional considerations

There are several additional aspects that users should consider during prompt engineering, such as approaches for multi-modal data types and formatting the output. Effectively integrating various data types and crafting precise prompts is crucial for maximizing the utility of models like GPT-5. For example, single cell RNA sequencing data can be transformed into detailed textual prompts that include gene markers and expression levels[91]. Similarly, biosensor monitoring signals can also be transformed into texts to enable the generation of personal health insights and exercise recommendations[92]. These prompts help the model to accurately perform multi-modal data analysis. Users should also consider the output format during prompt engineering, with the primary consideration being the difficulty of automatically parsing the response output. We therefore recommend instructing LLMs to generate a structured output, such as a JSON dictionary, to allow the result to be easily parsed into different answer sections (Box 1, Best Practice 8). In practice, users may also explore more anecdotal prompt refinements, such as focused examples, how context is placed or wrapped, or how multi-shot examples are organized, particularly during early iterative development.

## Fine-tuning

Although LLMs can solve many tasks using prompt engineering without explicit model modification, there are at least three situations where fine-tuning may be considered: (1) prompt engineering techniques like few-shot learning and RAG cannot sufficiently improve results, (2) high-quality training data is readily available in large scale, (3) the working prompt is too long to be feasible in terms of cost. (Box 2, Best Practice 9).

LLMs can be fully or partially fine-tuned. Full model fine-tuning updates all the parameters of an LLM, while parameter-efficient fine-tuning (PEFT) methods[93-95], such as Low-Rank Adaptation (LoRA)[93], that train much less parameters than the full-size LLM. In general, smaller and more specific data is suitable for PEFT to prevent overfitting[96], while larger and more diverse data is suitable for full-scale model fine-tuning[97]. PEFT greatly reduces the hardware requirements for fine-tuning, and is often competitive with full model fine-tuning methods and even outperform them in low-data environments. For example, quantized LoRA[94] was used to fine-tune LLMs for effective clinical text summarization using only thousands of training instances and one NVIDIA Quadro RTX 8000 GPU[21].

In summary, we suggest performing full-scale or parameter-efficient fine-tuning depending on computational resources and dataset features. Fine-tuning of LLMs can be done on a local server or through web-based graphical user interface provided by LLM and cloud vendors, which lowers the technical requirements but the implementation details might not be fully transparent. Similar to prompt engineering approaches, fine-tuned models also need to be evaluated on an independent test set to verify the performance improvement of training. More details on fine-tuning LLMs for medical use cases are discussed in relevant papers[98,99].

## Deployment considerations

### Regulatory compliance

Deploying LLMs in the biomedical domain requires adherence to privacy standards such as HIPAA and the General Data Protection Regulation[100] to protect patient data. When utilizing proprietary LLMs, it is essential to ensure that the platforms are HIPAA-compliant. Alternatively, processing data locally using an open-source model can enhance data safety[72]. For example, while the OpenAI API is not currently compliant with HIPAA, Azure services provide HIPAA-compliant access to OpenAI's models. Similarly, Anthropic provides HIPAA-certified API hosting for its Claude models. AI algorithms like LLMs are potentially regulated as medical devices, especially when used in clinical settings or for decision support. This adds an additional layer of regulatory scrutiny, requiring compliance with relevant standards such as those established by the US Food and Drug Administration (FDA) or other regulatory bodies overseeing medical device software. In the end, users must maintain the ethical and legal integrity of their deployment by carefully selecting protocols that align with compliance requirements and clinical standards (Box 2, Best Practice 10). For example, a hospital fine-tuning an LLM on secure local servers using fully de-identified patient notes, with access restricted to authorized internal staff, would be taking steps toward compliance with privacy regulations. Uploading identifiable patient records to public LLM APIs, on the other hand, would likely breach these regulations, especially when such APIs may use user inputs to train their models. Practitioners should carefully review the provider's user agreement before entering any patient-related information, and best practice is to use only commercial APIs that explicitly state HIPAA compliance or to deploy local models within the institution's secure infrastructure.

**System Robustness**

Users should evaluate potential biases in LLM's training data and algorithms to ensure fair and equitable outcomes[101]. Prior work has shown that even the most successful proprietary models can exhibit racial bias. Studies have shown that, when presented with identical patient profiles differing only by race, LLMs can yield varying predictions for treatment, cost, or outcome. Such differences can result in healthcare disparities during production. Thus, it is necessary to evaluate model fairness before deployment[102,103]. When examining data or algorithms is not viable, such as in the case of many proprietary models, users may still use existing benchmark datasets for evaluation[104,105]. This approach provides an idea of whether the models are fair or biased, and to what extent they exhibit bias.

**Costs**

When considering the costs associated with deploying large language models, it is important to distinguish between proprietary and open-source models. Proprietary LLMs require usage fees for each request made to the service provider. In the case of Google's Gemini 3.0 model, the pricing is $2 per one million input tokens for prompts less than 200k tokens, with additional charges for $12 per one million completion tokens (as of December 2025). For one thousand typical MIMIC-III[106] discharge note containing around four million tokens, processing would cost approximately $8 for prompt tokens, with additional costs depending on the response length generated by the model. Some proprietary model providers also offer customization services such as fine-tuning for an additional fee. Though proprietary models typically offer robust support, this pricing structure can cause delays in customization and updates. Utilizing services in this way does not guarantee access to the most advanced updates and limits customization, as new updates undergo extensive quality checks and alignments before companies release them. In contrast, open-source LLMs involve procurement costs for the necessary hardware, like GPUs, and ongoing costs related to maintenance. Resource requirements vary substantially depending on model size, quantization level, and expected throughput. For example, smaller instruction-tuned models (e.g., 7-8B parameters) can often be served efficiently on a single high-memory GPU (e.g., NVIDIA A100 40GB) or even on a CPU with reduced latency tolerance, whereas larger models (e.g., 70B parameters) typically require multi-GPU configurations or tensor-parallel serving. To improve computational efficiency and reduce hardware requirements, more recent models such as Qwen3[54] provide 8-bit floating-point (FP8) variants, in which model parameters are stored and computed using 8-bit floating-point representations rather than standard 16- or 32-bit floating-point formats, reducing GPU memory usage by approximately half compared with FP16. Cloud providers also offer GPU instances suitable for model hosting, allowing users to tradeoff between on-demand flexibility and long-term reserved capacity for cost efficiency. While the up-front costs are higher when running the model locally, these can be offset by the lower ongoing costs. In addition, an open-source setup can offer additional benefits like the ability to fine-tune the model to specific applications with protected data and more control over system responsiveness and updates. However, users should consider the potential for increased latency and reduced throughput during periods of high local demand. Local devices running LLMs might not match the speed and response time of large companies hosting LLMs.

**Post-deployment**

After the deployment of LLMs in healthcare, continuous monitoring is essential (Box 2, Best Practice 10). Users should ensure that LLM outputs are responsibly used as support tools, not as independent replacements for the judgment of healthcare practitioners. Effective training programs[107] are crucial to help healthcare professionals understand how to interpret and utilize these outputs while managing potential risks. Additionally, the successful integration of LLMs in medical practices demands active collaboration with patients and local communities. This involves leveraging engagement methods, such as community advisory boards and patient panels, to gather meaningful feedback and perspectives[108]. Such inclusive strategies help tailor LLM applications to the real-world diversity of patient experiences and enhance the effectiveness of these technologies in medical practice. Operationalizing LLMs further requires planning for model lifecycle changes, since the deprecation or replacement of production models may require re-implementation of prompts, re-validation of performance, and re-certification of safety and compliance.

## Conclusions

Large language models (LLMs) have the potential to revolutionize healthcare by enhancing clinical workflows, decision-making, and patient outcomes. However, their effective integration into medical practice requires a systematic and thoughtful approach. This tutorial provides a comprehensive framework for utilizing LLMs in medicine, emphasizing critical stages such as task formulation, model selection, prompt engineering, and deployment. By following these guidelines, healthcare professionals can maximize the benefits of LLMs while addressing challenges related to regulatory compliance, fairness, and cost. As AI continues to evolve, the careful application of these methods will be essential in ensuring that LLMs are used responsibly and effectively, ultimately improving the quality of care delivered to patients.

## Data Availability

Not applicable.

## Code Availability

Tutorial scripts are provided at https://github.com/ncbi-nlp/LLM-Medicine-Primer.

## Competing Interests Statement

The authors declare no competing interests.

## Acknowledgments

This research was supported by the Intramural Research Program of the National Institutes of Health (NIH). The contributions of the NIH author(s) are considered Works of the United States Government. Q.J. was also supported by the NIH Pathway to Independence Award K99LM014903. The findings and conclusions presented in this paper are those of the author(s) and do not necessarily reflect the views of the NIH or the U.S. Department of Health and Human Services.

**Box 1. Glossary**

1. **Large language model:** A large language model is a type of artificial intelligence designed to process and generate human-like text. The model is built using deep learning techniques and is trained on text corpora to perform tasks such as language translation, summarization, and question answering.
2. **Instance:** An instance refers to a specific example used for training or testing a model. Each instance typically includes input data and corresponding output, which the model is expected to predict or generate.
3. **Rationale:** Rationale is text output produced by a language model that provides additional context or reasoning for an output answer or value.
4. **Data modality:** A data modality is a category of data defined by how information is represented and acquired, such as clinical text, medical images, waveforms, genomic sequences, or structured electronic health records.
5. **Multi-modal comprehension**: Refers to the ability of models like GPT-5 to analyze and integrate diverse data types, including text, images, audio, and genomic information, for various applications such as medicine and research.
6. **Parameter:** A parameter is a variable within a large language model that is learned from training data and used to make predictions. The value of a parameter is often adjusted to minimize error during the training phase.
7. **Context Length/Window:** Context length/window refers to the number of tokens that a language model can receive as input. Larger context windows can increase the ability to perform in-context learning within a large language model prompt.
8. **Token**: A token is the smallest unit of information that a language model can process. Tokens often represent text information like single letters, spaces, sub-words, words, or phrases. According to empirical evidence, one token is about 0.8 of a word.
9. **Prompt engineering**: Prompt engineering involves crafting inputs or "prompts" that guide large language models to generate desired outputs without changing their parameters. Effective prompt design can significantly enhance the performance of LLMs.
10. **Few-shot learning**: Few-shot learning refers to the ability of a model to learn a new task from a very limited number of examples (shots). In the context of LLMs, this involves providing a few specific examples during the prompt to guide the model on how to handle similar tasks, improving its accuracy and adaptability.
11. **Retrieval-augmented generation**: Retrieval-augmented generation combines traditional language modeling with a retrieval component that searches for relevant information. This enhances the model's responses by grounding them in factual data.
12. **Tool learning**: Tool learning refers to the capability to recognize when external tools (e.g., search engines, calculators, databases, or code execution) are needed, select appropriate tools, and generate structured actions or calls to use them effectively.

13. **Application programming interface (API):** An API is a standardized protocol that enables software systems or models to communicate with external services, functions, or data sources by sending structured requests and receiving structured responses.
14. **Hallucination**: Hallucination refers to instances where the model generates incorrect or fabricated information, such as an incorrect factual claim or a non-existing citation.
15. **Chain-of-thought prompting**: Chain-of-thought prompting is a technique used with large language models to encourage step-by-step reasoning in their responses. By explicitly asking the model to think through the steps of a problem, it can provide more transparent and logically structured solutions.
16. **Temperature**: Temperature controls the randomness of the generated responses from large language models. A higher temperature increases diversity in the output, leading to more creative and varied responses, while a lower temperature produces more predictable and consistent outputs.
17. **Fine-tuning**: Fine-tuning is an approach to transfer learning in which pre-trained model parameters are further modified on a new dataset.
18. **Parameter-efficient fine-tuning (PEFT)**: PEFT updates only a small subset of model parameters or by introducing lightweight, trainable components, while keeping most model parameters fixed.
19. **HIPAA-compliance**: HIPAA-compliant systems adhere to HIPAA (Health Insurance Portability and Accountability Act) regulations, ensuring they meet the required standards to protect patient health information. HIPAA sets national standards for the protection of health information in the United States. It ensures the privacy and security of individually identifiable health information.

**Box 2. Best Practices**

1. Organize your task into one of the five categories: (1) **knowledge and reasoning**, (2) **summarization**, (3) **translation**, (4) **structurization**, and (5) **multi-modal data analysis**. Collect up to 100 test cases (instances) for evaluation with task-specific metrics. (Figure 2)
2. Ensure that the LLM usage is compliant to Health Insurance Portability and Accountability Act (**HIPAA**)[60] if working with sensitive data.
3. Choose an LLM that can process the **data modality** and has sufficient **context length** (length of the input prompt) used in the task. (Figure 3)
4. Select LLMs based on task-specific performance evaluated by experts in the literature. Scores on automatic evaluations can be used for the initial screening of LLMs (Figure 3)
5. Use one to five representative and diverse examples in **few-shot learning** to better specify the response style and handle edge cases. (Figure 4)
6. Always ask LLMs to "think-step-by-step" and generate the rationale before the answer for better explainability and improved performance. (**chain-of-thought prompting**[76]**,** Figure 4)
7. Use **retrieval-augmented generation (RAG)**[81] or tool learning if knowledge or domain utilities are needed and to make evidence-based and up-to-date generation. (Figure 4)
8. Use a **temperature** of 0 and **JSON formatting** to generate reproducible and structured outputs that can be easily parsed. (Figure 4)
9. Consider **fine-tuning** when prompt engineering fails to reach the desired performance, the working prompt is too costly, or abundant training data is readily available. (Figure 4)
10. Safeguard against potential risks by monitoring **regulatory compliance, system robustness**, and **cost** when deploying LLMs in the biomedical domain[69].

**Table 1**. Characteristics of different LLMs, sorted by the highest-reported MedQA-USMLE[65] (4 options) score. T: text; I: image; V: video; A: audio. The GPT-5 series include GPT-5, GPT-5 mini, GPT-5 nano, and GPT-5 pro. The Claude 4.5 series include Claude 4.5 Opus, Claude 4.5 Sonnet, and Claude 4.5 Haiku. The GPT-4 series includes GPT-4, GPT-4-turbo, and GPT-4o. The GPT-3.5 series includes Codex and GPT-3.5-turbo.

| LLM | Weights | Size | Interface | Modality | Context | MedQA |
|---|---|---|---|---|---|---|
| **GPT-5** | Closed | NA | Web, API | T, I | 400k | 95.8%[109] |
| **o1-preview** | Closed | NA | Web, API | T | 128k | 94.9%[58] |
| **Gemini 3.0 Pro** | Closed | NA | Web, API | T, I, V, A | 1M | 94.6%[110] |
| **o3-mini** | Closed | NA | Web, API | T | 200k | 92.7%[59] |
| **Gemini 2.5 Pro** | Closed | NA | Web, API | T, I, V, A | 1M | 92.6%[111] |
| **DeepSeek-R1** | Open | 671B | Web, API, Local | T | 128k | 92.0%[59] |
| **Claude 4.5** | Closed | NA | Web, API | T, I | 200k | 91.4%[110] |
| **Med-Gemini** | Closed | NA | Web, API | T, I, V, A | 1M, 2M | 91.1%[9] |
| **GPT-4** | Closed | NA | Web, API | T, I | 8k, 32k, 128k | 90.2%[14] |
| **Llama 3.1** | Open | 8B, 70B, 405B | API, Local | T | 128k | 88.2%[112] |
| **Qwen3** | Open | 0.6B, 1.7B, 4B, 8B, 14B, 30B, 32B, 235B | Web, API, Local | T | 32k, 262k | 87.4%[113] |
| **Med-PaLM 2** | Closed | NA | API | T | 8k | 86.5%[34] |
| **Llama 3** | Open | 8B, 70B, 405B | API, Local | T | 8k | 80.9%[81] |

| **GPT-3.5** | Closed | NA | Web, API | T | 4k, 16k | 68.7%[114] |
|---|---|---|---|---|---|---|
| **Med-PaLM** | Closed | 540B | API | T | 8k | 67.6%[15] |
| **Gemini 1.0** | Closed | NA | Web, API | T, I, V | 32k | 67.0%[115] |
| **Mixtral** | Open | 8x7B | API, Local | T | 32k | 64.1%[81] |
| **Mistral** | Open | 7B | API, Local | T | 8k, 32k | 59.6%[81] |
| **Llama 2** | Open | 7B, 70B | API, Local | T | 4k | 47.8%[81] |

**Table 2**. Characteristics of different approaches to use large language models in medicine.

| **Approach** | **Requirements** | **Pros** | **Cons** | **Examples** |
|---|---|---|---|---|
| **Few-shot learning** | Several exemplars | 1. Dealing with edge cases;<br>2. Specifying expected styles | Exemplars might introduce biases | MedPrompt[14] |
| **Tool learning** | Application programming interfaces | Providing domain functionalities | Relies on the curation of tools | GeneGPT[86], GeneAgent[87], EHRAgent[89], ChemCrow[116] |
| **Chain-of-thought prompting** | Additional prompt text ("Let's think step-by-step.") | 1. Providing explanations;<br>2. Improving performance | Hard to parse (mitigated by structured output) | MedPrompt[14] |
| **Retrieval-augmented generation** | A knowledge base or document collection | 1. Providing up-to-date knowledge;<br>2. Reducing hallucinations | Depends on the quality of the retrieved documents | Almanac[117], MedRAG[81] |
| **Fine-tuning** | Data annotations and compute | 1. Improving performance<br>2. Shorten the prompt | Full-scale fine-tuning can be costly and resource intensive | MEDITRON[57], PMC-LLaMA[56] |

**Table 3**. Representative case studies of utilizing large language models in medicine.

| **Study** | **Task Formulation** | **LLM(s) Selection** | **Technique** | **Evaluation** |
|---|---|---|---|---|
| Van Veen et al.[21] | Summarization | FLAN-T5[97], Llama, and etc. | Few-shot learning, fine-tuning | Automatic and manual evaluation of summaries |
| Singhal et al.[15] | Knowledge and Reasoning | Med-PaLM | Few-shot learning, CoT prompting, fine-tuning | MCQ evaluation and manual evaluation of answers |
| Wang et al.[48] | Structurization | Llama | Fine-tuning | Automatic classification evaluation and manual error analysis |
| Mirza et al.[47] | Translation | GPT-4 | Direct prompting | Manual evaluation of clinical translation by domain experts |
| Zhang et al.[118] | Multi-modality | BiomedGPT[118] | Fine-tuning | MCQ evaluation and manual evaluation of visual tasks |